\documentclass[9pt]{article}

\usepackage{spconf,amsmath,graphicx}

\usepackage{amsmath,amsfonts,bm}

\def\eqref#1{equation~\ref{#1}}

\def\1{\bm{1}}

\DeclareMathAlphabet{\mathsfit}{\encodingdefault}{\sfdefault}{m}{sl}
\SetMathAlphabet{\mathsfit}{bold}{\encodingdefault}{\sfdefault}{bx}{n}

\usepackage{xcolor}
\usepackage{hyperref}
\usepackage{url}
\usepackage{adjustbox}
\usepackage{float}
\usepackage{multirow}

\newcommand\copyrighttext{%
    \small \begin{center} \color{red} \textcopyright\,2024 IEEE. Personal use of this material is permitted. Permission from IEEE must be obtained for all other uses, in any current or future media, including reprinting/republishing this material for advertising or promotional purposes, creating new collective works, for resale or redistribution to servers or lists, or reuse of any copyrighted component of this work in other works. \end{center}}

\title{\vspace{-5em} \copyrighttext \vspace{1em} \large DIffSTOCK: Probabilistic RElational Stock Market Predictions using DIFFUSION MODELS}
\name{Divyanshu Daiya$^{1}$, Monika Yadav$^{2}$, Harshit Singh Rao$^{3}$}
\address{
 $^1$Department of Computer Science, 
Purdue University\\
 $^2$Thomas Lord Department of Computer Science, University of Southern California, $^3$NexusQuant AI\\
\small $^1$divyanshu@purdue.edu, \small $^2$monikaya@usc.edu,  \small $^3$harshitrao@nexusquant.tech}

\begin{document}
%
\maketitle
\begin{abstract}
In this work, we propose an approach to generalize denoising diffusion probabilistic models for stock market predictions and portfolio management. Present works have demonstrated the efficacy of modeling interstock relations for market time-series forecasting and utilized Graph-based learning models for value prediction and portfolio management. Though convincing, these deterministic approaches still fall short of handling uncertainties i.e., due to the low signal-to-noise ratio of the financial data, it is quite challenging to learn effective deterministic models. Since the probabilistic methods have shown to effectively emulate higher uncertainties for time-series predictions. To this end, we showcase effective utilisation of Denoising Diffusion Probabilistic Models (DDPM), to develop an architecture for providing better market predictions conditioned on the historical financial indicators and inter-stock relations. Additionally, we also provide a novel deterministic architecture MaTCHS which uses Masked Relational Transformer(MRT) to exploit inter-stock relations along with historical stock features. We demonstrate that our model achieves SOTA performance for movement predication and Portfolio management.\\

\end{abstract}
\vspace{-15pt}

\section{Introduction}
Stock price prediction is an age-old intrigue for investors due to its potential dividends and the inherent challenges it presents due to market volatility and its stochastic nature \cite{akita2016deep, chen2015lstm}. Modern advancements in Deep Learning now empower researchers to employ a multitude of modalities such as historical stock trends, news, social media, and financial reports in market prediction models \cite{ding2015deep, vargas2017deep}. There has been a concerted effort in modeling inter-stock dependencies, revealing that stocks associated with top-tier positions, or those in similar sectors, often show correlated trends, contributing to significant improvements in market predictions \cite{sawhney2021stock, feng2019temporal}.

Historically, time series prediction relied heavily on state space framework-based statistical models such as ARIMA and exponential smoothing \cite{hyndman2008forecasting, box2015time}. However, while pure machine learning approaches have been explored, they often didn't surpass statistical models due to issues like overfitting and non-stationarity \cite{bandara2020forecasting, makridakis2018statistical}. In recent years, the spotlight has shifted towards the diffusion model for probabilistic time series forecasting, marking state-of-the-art performances. Notably, the TimeGrad model \cite{rasul2021autoregressive} and CSDI model \cite{tashiro2021csdi} have showcased the potency of the diffusion model for optimizing forecasting.

Yet, a challenge persists in the realm of stock prediction: the low signal-to-noise ratio inherent in stock data. Such noise can hinder machine learning models' effectiveness, affecting the accuracy of latent factors \cite{israel2020can}. While the integration of multi-modal data helps bridge the gap, deterministic methods often grapple with the uncertainties introduced over time. This has led to the rise of probabilistic models, notably the Denoising Diffusion Probabilistic Models (DDPM) \cite{ho2020denoising}, which transform noise into predictions using a denoising process conditioned on historical readings. However, their limitation lies in solely modeling temporal dependencies, neglecting spatial correlations between stock nodes \cite{rasul2021autoregressive, tashiro2021csdi}.

To surmount these challenges, we propose a novel framework that synergizes DDPMs with relational market data, encapsulating the spatio-temporal strengths of deterministic models and the uncertainty handling of DDPMs. Our MaTCHS architecture, derived from our preceding work \cite{daiya2021stock}, employs Transformer Encoders with Masked attention heads, encapsulating both temporal dynamics and spatial correlations.

For a detailed exploration of deterministic models, readers are encouraged to refer to \cite{daiya2021stock,9053479}.

\section{Model}
\label{sec:pagestyle}
\subsection{General Market Prediction Task}
We are interested in the problem of predicting future stock prices, and we are provided with $P$ financial indicators for the last $L$ days for every stock. We have a total of $N$ stocks, and which gives a set of financial and social indicators represented as, $f_{N,L}=(f'^{1}_{P,L},f'^{2}_{P,L}...,f'^{N}_{P,L})$ here $f'^{m}_{P,L} \in \mathbb{R}^{P\times L}$.\\
In addition, we have a relation matrix  $C \in \mathbb{R}^{N\times N\times G}$, which specifies connections between $N$ stocks over $G$ different relations, $C(i,j,k)=\{$1 if i and j are connected by relation k, else 0$\}$\cite{feng2019temporal}. We can model our relation matrix as a graph $H=(\mathcal N, \mathcal E)$, with ${\mathcal E}\in \mathbb{R}^{G}$ is a hyperedge.\\
Given historical financial signals $f_{N,L}$ we can represent each node as a stock in the graph $H$. Now, given this information we seek to predict the stock value at time step $T_{observed}+1$ to $T$ for all the $N$ stocks, i.e. $x^{(T+1,N)}$. So, we can formulate our problem as $p(x^{(T+1,N)}|f_{N,L}, H)$ and we seek to learn some $F:(f_{N,L}, H) \rightarrow x^{(T+1,N)}$. 

\subsection{Conditional Diffusion Model}

We build on the work by \cite{tashiro2021csdi, wen2023diffstg}, to develop a conditional diffusion model for market prediction. We can start by following a obvious approach using the stock trend history $f_{t,L}$ and relational data between stocks $H$ as the condition in the reverse process. So, we can write our conditioned reverse diffusion process as,
\begin{align}
& p_{\theta}(x^{(T+1,N)}_{0:K}|f_{N,L}, H) = p(x^{(T+1,N)}_{K}) \nonumber \\
& \times \prod_{k=K}^{1}p_\theta(x^{(T+1,N)}_{k-1}|x^{(T+1,N)}_{k}, f_{N,L}, H). 
\label{eq:condition_reverse_process}
\end{align}
Here we can note that $x^{(T+1,N)}$ can be thought of as being sampled from the same distribution as $f_{N,L}$ due to the high trend correlation between consecutive time-stamps.
To better utilize this association, we modify the equation to predict $f_{N,L+1}$ i.e. $f_{N,L}$ along with future time steps as demonstrated by \cite{wen2023diffstg}. This formulation provides a unifying approach that combines historical reconstruction and future estimation. By predicting $f_{N,L+1}$, we can use historical data to model the distribution of data comprehensively.
\begin{align}
& p_{\theta}((f_{N,L+1})_{0:K}|f_{N,L}, H) =  p((f_{N,L+1})_{K}) \nonumber \\
& \times \prod_{k=K}^{1}p_\theta((f_{N,L+1})_{k-1}|(f_{N,L+1})_{k}, f_{N,L}, H). 
\label{eq:condition_reverse_process}
\end{align}
with training objective,
\begin{equation}
\small
\mathop{\min} _\theta \mathcal{L}(\theta)=
\min _\theta \mathbb{E}_{(f_{N,L+1})_{0}, \epsilon} \left\|\epsilon-\epsilon_\theta\left((f_{N,L+1})_k, k | f_{N,L}, H\right)\right\|_2^2.
\label{eq:masked_condition_loss}
\end{equation}
Here, $\epsilon_\theta$ is our MaTCHS model, the denoising model. The denoising function $\epsilon_\theta$ estimates the noise vector $\epsilon$ that was added to its noisy input $(f_{N,L+1})_k$. Detailed formulation is similar to works of \cite{rasul2021autoregressive,tashiro2021csdi,ho2020denoising}. \footnote{It is to be noted that at all the instance where we have used $T+1$ and $L+1$ it can be generalised to $T+t'$ and $L+l'$ i.e. any number of future time steps $t'$ or $l'$ can be predicted.}

\begin{figure*}[ht!]
\centering
  \includegraphics[width=0.9\textwidth]{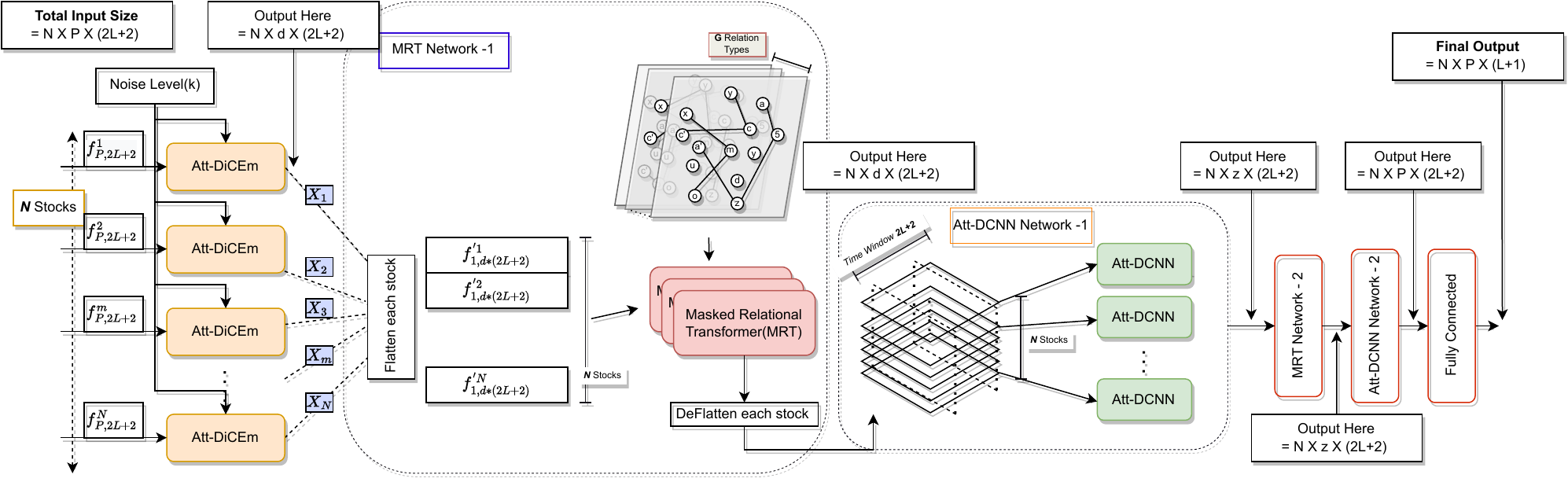}
\caption{$MaTCHS$ Denoising Model: Masked Transformer and Convoutional network for Hypergraph relation based Stock time-series generation.}
    \label{match}
\vspace{-10pt}
\end{figure*}

\subsection{Adaptive Noise for Financial Series Diffusion}
Stock market dynamics are inherently stochastic and diverse, exhibiting various unique patterns like spikes, trends, and declines at different time points. To adequately model and capture these diverse dynamics, it's imperative for the Difffusion process to model inherent volatility at different time points and the collective behaviors of groups of stocks. 

\textbf{Modeling Time Series Variance:}
One way to assess the intrinsic unpredictability or volatility of the stock market is by evaluating the local variance at each time point. This local variance can be thought of as an indicator of how sensitive a stock might be to broader market fluctuations. The formula is given by,
$ v(t) = \frac{\sum_{i=t-w}^{t+w} (f_{N,t} - f_{N,i})^2}{2w+1}$,
Where the normalization of this variance is represented as,
$v_{norm}(t) = \frac{v(t)}{\max_{\tau \in T} v(\tau)}.$ 

\textbf{Modeling Intra-Cluster Dynamics:}
Groups of stocks often exhibit collective behaviors, especially if they belong to the same sector or are influenced by similar macroeconomic factors. To model such collective behaviors or intra-cluster dynamics, we employ Dynamic Time Warping (DTW)\cite{assent2009anticipatory} method. DTW provides a measure of similarity between time series of individual stocks. Thus, the intra-cluster DTW distance for a stock within a cluster can be expressed as:
\begin{align*}
\small
DTW_{\text{intra}}(f_{N,i}, C) = \text{DTW}\left(\frac{1}{|C|-1}\sum_{f_{N,j} \in C, i \neq j} f_{N,i}, f_{N,j}\right)
\end{align*}
The influence of a stock's time series in relation to its cluster is then given by, $I_{intra}(f_{N,i}, C) = \frac{1}{1 + DTW_{intra}(f_{N,i}, C)}$.

\textbf{Integrating Volatility and Cluster Dynamics:}
To obtain a comprehensive understanding of a stock's behavior, both its individual volatility and its intra-cluster dynamics should be taken into account. By integrating these, we get:
\begin{align*}
v_{score,intra}(t, f_{N,i}, C) = \alpha \times v_{norm}(t) + (1-\alpha) \\ \times I_{intra}(f_{N,i}, C) \times v_{norm}(t, C),
\end{align*}

This integrated score provides a unified metric, represented as \( I(t) = {v_{score,intra}}_{norm}(t, C) \), that captures the overall significance of a given time point. Based on this metric, noise can be adaptively applied to market signal :
\begin{align}
\small
\begin{split}
&q({(f_{N,L+1})}_{k+1}|{(f_{N,L+1})}_{k}) = \\
&N({(f_{N,L+1})}_{k+1}; \sqrt{(1-\beta_t)}{(f_{N,L+1})}_{k}, \beta_t I).
\end{split}
\end{align}
This ensures that significant time points and relation patterns are emphasized during Diffusion process by prioritizing learning these trends during denoising process and enabling model in making better future trend prediction.

\subsection{MaTCHS(Denoising Model)}

Introducing \textbf{MaTCHS}, our architecture fuses a \textbf{Ma}sked \textbf{T}ransformer with a \textbf{C}onvoutional network to predict stock prices leveraging \textbf{H}ypergraph relations. It consists of two main segments. The first, Att-DiCEm, focuses on temporal feature extraction from financial indicators using the Att-DCNN approach\cite{9053479}. The second emphasizes understanding the relationship between different stocks for price prediction. After the "Masked Relational Transformer(MRT)" provides relational information, this is further enhanced temporally with another Att-DCNN layer. The model concludes by producing a financial time-series for each stock spanning past and future time-stamps. The input to our denoising network will be concatenated $(f_{N,L+1})_k$, and $f_{N,L+1}$, the conditional for the diffusion model with mask at the future time-steps. So the concatenated input takes the form of $f_{N,2(L+1)}$.

\textbf{Att-DiCEm}
Building upon our previous work\cite{daiya2021stock}, the event pipeline has been excluded, emphasizing the financial segment. We've adapted this segment to generate a time series across 2L+1 time steps with "d" time variables. Contrasting the original output of $\mathbb{R}^{(2L+1)\times1}$, the new format is $\mathbb{R}^{(2L+1)\times d}$. Noise level positional encoding, $k \in [1,K]$, is implemented using a transformer positional embedding\cite{vaswani2017attention} as, 
$
\small
\mathbf{ne}(k)=[\dots, \cos(k / r^{-2s/D}), \sin(k / r^{-2s/D}), \dots]^{\mathrm{T}},
$
with the embeddings added post the initial DC-CNN layer in each Att-DiCEm unit.
\vspace{-10pt}
\subsubsection{Masked Relational Transformer (MRT)}
\label{transformer-layer}

As described in~ \cite{vaswani2017attention}, in the Self-Attention Network  an attention function maps a query and a set of key-value pairs to an output as, $
\mathcal{A}(Q,K,V) = \mathcal{S}(Q,K)V$,
with $\ \mathcal{S}(Q,K)=\Bigg[\frac{\text{exp}\big(Q_{i}K_{j}^{T}/\sqrt{d_{k}}\big)}{\sum_{k}\text{exp}\big(Q_{i}K_{k}^{T}/\sqrt{d_{k}}\big)}\Bigg]$. Here, queries $Q$, keys $K$, and values $V\in\mathbb{R}^{T\times d_{k}}$ are matrices. A representation sequence is given as $H^{l}\in \mathbb{R}^{T\times d}$ in the $l$-th layer, $H^{l}=\big[A^{1},\cdots,A^{I}\big] W_H $, $i$ denotes the attention head and $d$ is the hidden size.\\
By analyzing the self-attention function, we can model it as a graph. The relation from Node $j$ to $i$ can be described by, $rel(\text{i to j}) = q_i^\mathrm{T}k_j$. A modified attention function, incorporating masking to make the model operate on specific graph nodes, is described as, $\mathcal{A}_{M}(Q,K,V) =\ {S}_{M}(Q,K)V$
with, $\mathcal{S}_{M}(Q,K)=\Bigg[\frac{M_{i,j}\text{exp}\big(Q_{i}K_{j}^{T}/\sqrt{d_{k}}\big)}{\sum_{k}M_{i,k}\text{exp}\big(Q_{i}K_{k}^{T}/\sqrt{d_{k}}\big)}\Bigg]$, where $M\in\mathbb{R}^{T\times T},M_{i,j}\in[0,1]$ can be a static or dynamic mask matrix.\\
Now to model hyper-graph structure one approach would be of aggregating all relation types into a single matrix, but that might overlook specific relation nuances~\cite{vaswani2017attention}. Given that multi-head attention mechanisms sometimes capture redundant features~\cite{michel2019sixteen,voita2019analyzing,lu202012}, we propose the Masked Relational Attention Networks (MRAN). It uses separate attention heads for each relation, $        \mathcal{A}^{i}_{M^i}(Q^{i},K^{i},V^{i}) = \mathcal{S}^{i}_{M^i}(Q^{i},K^{i})V^{i}$ with$, 
        \mathcal{S}^{i}_{M^i}(Q^{i},K^{i})=\Bigg[\frac{M^{i}_{i,j}\text{exp}\big(Q^{i}_{i}K_{j}^{iT}/\sqrt{d_{k}}\big)}{\sum_{k}M^{i}_{i,k}\text{exp}\big(Q^{i}_{i}K_{k}^{iT}/\sqrt{d_{k}}\big)}\Bigg]
$, such that, ${A}^{i}_{M}$ is Attention calculated for $i^{th}$ attention head which uses $i^{th}$ relation mask $M^i$ i.e. $C(:,:,i)$.\\
Due to computational concerns, we restrict attention heads in our transformer to 12, grouping similar relations together and using aggregated relations as masks. So, Masked Relational Transformer (MRT), divides the total $G$ relations among 12 attention heads. We also employ additional 4 unmasked attention heads for capturing potential evolving relations between stocks. So, we employ grouped stock-relations C as our second conditional. Given output $X={X_1,\dots,X_N}$ from the Att-DiCEm layer, we first flatten it to $X'_m \in \mathbb{R}^{d(2L+1) \times 1}$. MRT uses $X'_t$ as input for Transformer Encoder Layer. The output, $X''_t$, is obtained by de-flattening each stock in $\mathrm{X''_t}$ to match MRT layer input dimensions. Note that these Transformer Encoders model spatial relations among stocks, not temporal domains.\\
For training, we retained preprocessing from \cite{daiya2021stock} and used the diffusion model hyperparameters outlined by \cite{tashiro2021csdi}. We varied $\beta_k$ capped at $[0, 1, 0.2]$ and diffusion steps $K$ from [20, 50, 100, 200]. Using a batch size of 16 and a decaying learning rate initialized at $1e-4$, training spanned 100 epochs. Two transformer encoders in each MRT Network were used with a prediction window of 3. Evaluation for StockNet was centered on the L+1 timestamp. Training with 200 diffusion steps for 100 epochs on the A100 GPU took 6 hours. Optimal values on StockNet were $K=100$ and $\beta_k=0.2$.

\section{Evaluation}
\subsection{Datasets and Baselines}
For an extensive evaluation of our model, we plan to evaluate it on four datasets from $U S$ Stock Market spanning over 6 years. We test on NASDAQ \cite{feng2019temporal}, NYSE \cite{feng2019temporal} and StockNet \cite{xu-cohen-2018-stock}. We follow the same train test partitions as in original works \cite{feng2019temporal, xu-cohen-2018-stock}. 
We follow the approach described by \cite{feng2019temporal} for populating relation matrix $C$ for all the datasets. We compare our model with top performing models, STHGCN \cite{Sawhney2020Spatiotemporal}, GCN \cite{Chen2018Incorporating}, GCN20 \cite{Kim2019HATS},  RSR \cite{feng2019temporal}, HATS \cite{Kim2019HATS}. For Stock Movement Prediction, we'll use accuracy and MCC. For Portfolio Management, we'll compare our cumulative return (IRR) with other trading models and use Sharpe Ratio to measure risk-adjusted returns. For probabilistic models (DDPM), we'll use CRPS to assess distribution compatibility. For portfolio management using predicted prices we will adopt a daily buy-hold-sell strategy as described by \cite{sawhney2021stock}.
\vspace{-15pt}
\begin{table}[h]
\caption{Evaluation Results over StockNet Dataset\cite{xu-cohen-2018-stock}}
\centering
\resizebox{0.47\textwidth}{!}{%
\begin{tabular}{llrrr}
\hline
 & Model & ${F 1} \uparrow$ & Accuracy $\uparrow$ & MCC $\uparrow$ \\
\hline
 & RAND & $0.502 \pm 8 e-4$ & $0.509 \pm 8 e-4$ & $-0.002 \pm 1 e-3$ \\
TA & ARIMA - \cite{brown2004smoothing} & $0.513 \pm 1 e-3$ & $0.514 \pm 1 e-3$ & $-0.021 \pm 2 e-3$ \\
 & - \cite{selvin2017stock} & $0.529 \pm 5 e-2$ & $0.530 \pm 5 e-2$ & $-0.004 \pm 7 e-2$ \\
\hline
 & RandForest - \cite{pagolu2016sentiment} & $0.527 \pm 2 e-3$ & $0.531 \pm 2 e-3$ & $0.013 \pm 4 e-3$ \\
 & TSLDA - \cite{nguyen2015topic} & $0.539 \pm 6 e-3$ & $0.541 \pm 6 e-3$ & $0.065 \pm 7 e-3$ \\
 & HAN - \cite{10.1145/3159652.3159690} & $0.572 \pm 4 e-3$ & $0.576 \pm 4 e-3$ & $0.052 \pm 5 e-3$ \\
 & StockNet - TechnicalAnalyst - \cite{xu-cohen-2018-stock} & $0.546 \pm-$ & $0.550 \pm-$ & $0.017 \pm-$ \\
 & StockNet - FundamentalAnalyst - \cite{xu-cohen-2018-stock} & $0.572 \pm-$ & $0.582 \pm-$ & $0.072 \pm-$ \\
 & StockNet - IndependentAnalyst - \cite{xu-cohen-2018-stock} & $0.573 \pm-$ & $0.575 \pm-$ & $0.037 \pm-$ \\
FA & StockNet - DiscriminativeAnalyst - \cite{xu-cohen-2018-stock} & $0.559 \pm-$ & $0.562 \pm-$ & $0.056 \pm-$ \\
 & StockNet - HedgeFundAnalyst - \cite{xu-cohen-2018-stock} & $0.575 \pm-$ & $0.582 \pm-$ & $0.081 \pm-$ \\
 & GCN\cite{Chen2018Incorporating} & $0.530 \pm 7 e-3$ & $0.532 \pm 7 e-3$ & $0.093 \pm 9 e-3$ \\
 & HATS - \cite{Kim2019HATS} & $0.560 \pm 2 e-3$ & $0.562 \pm 2 e-3$ & $0.117 \pm 6 e-3$ \\
 & Adversarial LSTM - \cite{feng2019temporal} & $0.570 \pm-$ & $0.572 \pm-$ & $0.148 \pm-$ \\
 & MAN-SF - \cite{sawhney-etal-2020-deep} & ${0 . 6 0 5} \pm {2 e}-{4}$ & ${0 . 6 0 8} \pm {2 e -  { 4 }}$ & ${0 . 1 9 5} \pm {6 e - 4}$ \\
 & STHGCN - \cite{sawhney2021stock} & ${0 . 6 09} \pm {2 e}-{4}$ & ${0 . 6 13} \pm {2 e -  { 4 }}$ & ${0 . 198} \pm {6 e - 4}$ \\
& \textbf{MaTCHS (This work)} - AttDiCEm i.e. without relations & $\mathbf{0 .568} \pm \mathbf{2 e}-\mathbf{3}$ & $\mathbf{0 . 572} \pm \mathbf{2 e - \mathbf { 3 }}$ & $\mathbf{0 . 168} \pm \mathbf{6 e - 3}$ \\
& \textbf{MaTCHS (This work)} - Aggregated relations & $\mathbf{0 .585} \pm \mathbf{2 e}-\mathbf{3}$ & $\mathbf{0 . 587} \pm \mathbf{2 e - \mathbf { 3 }}$ & $\mathbf{0 . 175} \pm \mathbf{6 e - 3}$ \\  
& \textbf{MaTCHS (This work)} & $\mathbf{0 . 6 1 1} \pm \mathbf{2 e}-\mathbf{3}$ & $\mathbf{0 . 6 12} \pm \mathbf{2 e - \mathbf { 3 }}$ & $\mathbf{0 . 206} \pm \mathbf{6 e - 3}$ \\
& \textbf{MaTCHS (with Diffusion w/o Adap. Noise)} & $\mathbf{0 . 6 23} \pm \mathbf{2 e}-\mathbf{3}$ & $\mathbf{0 . 621} \pm \mathbf{2 e - \mathbf { 3 }}$ & $\mathbf{0 . 214} \pm \mathbf{6 e - 3}$ \\
& \textbf{MaTCHS (with Diffusion)} & $\mathbf{0 . 6 3 1} \pm \mathbf{2 e}-\mathbf{3}$ & $\mathbf{0 . 6 34} \pm \mathbf{2 e - \mathbf { 3 }}$ & $\mathbf{0 . 225} \pm \mathbf{6 e - 3}$ \\
\hline
\end{tabular}}
\end{table}
\vspace{-20pt}
\begin{table}[h]
\begin{center}
\caption{Evaluation Results over NASDAQ and NYSE Dataset(2 decimal places disp.) \cite{feng2019temporal}}
\scalebox{0.55}{
\begin{tabular}{lcccc}
\hline
Model & \multicolumn{2}{c}{NYSE} & \multicolumn{2}{c}{NASDAQ} \\
\cline { 2 - 5 }
 & SR@5 & IRR@5 & SR@5 & IRR@5 \\
\hline
ARIMA \cite{brown2004smoothing} & $0.33 \pm 3 e^{-3}$ & $0.10 \pm 5 e^{-3}$ & $0.55 \pm 1 e^{-3}$ & $0.10 \pm 6 e^{-3}$ \\
A-LSTM \cite{feng2018enhancing} & $0.81 \pm 4 e^{-3}$ & $0.14 \pm 7 e^{-3}$ & $0.97 \pm 5 e^{-3}$ & $0.23 \pm 3 e^{-3}$ \\
GCN \cite{Chen2018Incorporating} & $0.70 \pm 3 e^{-3}$ & $0.10 \pm 6 e^{-3}$ & $0.75 \pm 4 e^{-3}$ & $0.13 \pm 1 e^{-3}$ \\
HATS \cite{Kim2019HATS} & $0.73 \pm 5 e^{-3}$ & $0.12 \pm 2 e^{-3}$ & $0.80 \pm 6 e^{-3}$ & $0.15 \pm 7 e^{-3}$ \\
\hline
DQN \cite{carta2021multi} & $0.72 \pm 5 e^{-3}$ & $0.12 \pm 4 e^{-3}$ & $0.93 \pm 5 e^{-3}$ & $0.20 \pm 6 e^{-3}$ \\
iRDPG \cite{liu2020adaptive} & $0.85 \pm 7 e^{-3}$ & $0.18 \pm 3 e^{-3}$ & $1.32 \pm 5 e^{-3}$ & $0.28 \pm 4 e^{-3}$ \\
\hline
Rank LSTM \cite{bao2017deep} & $0.79 \pm 1 e^{-3}$ & $0.12 \pm 6 e^{-3}$ & $0.95 \pm 4 e^{-3}$ & $0.22 \pm 2 e^{-3}$ \\
GCN \cite{Chen2018Incorporating} & $0.72 \pm 7 e^{-3}$ & $0.16 \pm 3 e^{-3}$ & $0.46 \pm 4 e^{-3}$ & $0.13 \pm 5 e^{-3}$ \\
RSR-E \cite{feng2019temporal} & $0.88 \pm 6 e^{-3}$ & $0.20 \pm 3 e^{-3}$ & $1.12 \pm 5 e^{-3}$ & $0.26 \pm 4 e^{-3}$ \\
RSR-I \cite{feng2019temporal} & $0.95 \pm 1 e^{-3}$ & $0.21 \pm 3 e^{-3}$ & $1.34 \pm 6 e^{-3}$ & $0.39 \pm 5 e^{-3}$ \\
STHAN-SR \cite{sawhney2021stock} & $1 . 1 0 \pm \pm e^{-3}$ & $0 . 2 5 5 \pm e^{-3}$ & $1 . 4 0 \pm 7 e^{-3}$ & $0 . 4 4 \pm 1 e^{-2}$ \\
\hline
\textbf{MaTCHS} & $\mathbf{1.13} \pm \pm e^{-3}$ & $\mathbf{0 . 26} 7 \pm e^{-3}$ & $\mathbf{1 . 45} \pm 7 e^{-3}$ & $\mathbf{0 . 4 5} \pm 1 e^{-2}$ \\

MaTCHS(Agg) & $\mathbf{0.97} \pm \pm e^{-3}$ & $\mathbf{0 . 2 2} 1 \pm e^{-3}$ & $\mathbf{1 . 34} \pm 7 e^{-3}$ & $\mathbf{0 . 4 0} \pm 1 e^{-2}$ \\
MaTCHS(16) & $\mathbf{1.14} \pm \pm e^{-3}$ & $\mathbf{0 . 27} 0 \pm e^{-3}$ & $\mathbf{1 . 46} \pm 7 e^{-3}$ & $\mathbf{0 . 4 6} \pm 1 e^{-2}$ \\
\textbf{MaTCHS with Diffusion w/o Adap. Noise} & $\mathbf{1.15} \pm \pm e^{-3}$ & $\mathbf{0 . 27} 4 \pm e^{-3}$ & $\mathbf{1 . 48} \pm 7 e^{-3}$ & $\mathbf{0 . 4 6} \pm 1 e^{-2}$ \\
\textbf{MaTCHS with Diffusion} & $\mathbf{1.18} \pm \pm e^{-3}$ & $\mathbf{0 . 28} 5 \pm e^{-3}$ & $\mathbf{1 . 52} \pm 7 e^{-3}$ & $\mathbf{0 . 4 8} \pm 1 e^{-2}$ \\
\hline
\% Improv. (SOTA w.r.t. STHAN-SR) & $7.92$ & $9.81$ & $6.18$ & $8.07$ \\
\hline
\hline
\end{tabular}
}\end{center}
\label{table1}
\end{table}
\vspace{-25pt}

\begin{table}[h]
\centering
\small
\caption{\centering {Comparison with other Diffusion Models} }
\setlength\tabcolsep{6pt} 
\scalebox{0.64}{
\begin{tabular}{c|cccc}
\hline
Diffusion Model & \multicolumn{4}{|c|}{ StockNet }\\
\cline{2-5}
 & F1 & Accuracy & MCC & CRPS \\
\hline \hline
CSDI \cite{tashiro2021csdi}  &  $\mathbf{0 . 582} \pm \mathbf{2 e}-\mathbf{3}$ & $\mathbf{0 . 586} \pm \mathbf{2 e - \mathbf { 3 }}$ & $\mathbf{0 . 170} \pm \mathbf{6 e - 3}$ & 0.092\\  \hline
TimeGrad \cite{rasul2021autoregressive} &  $\mathbf{0 . 596} \pm \mathbf{2 e}-\mathbf{3}$ & $\mathbf{0 . 598} \pm \mathbf{2 e - \mathbf { 3 }}$ & $\mathbf{0 . 177} \pm \mathbf{6 e - 3}$ & 0.076\\ \hline
MaTCHS with Diffusion (ours)  &  $\mathbf{0 . 631} \pm \mathbf{2 e}-\mathbf{3}$ & $\mathbf{0 . 634} \pm \mathbf{2 e - \mathbf { 3 }}$ & $\mathbf{0 . 225} \pm \mathbf{6 e - 3}$ & 0.049\\
\end{tabular}
}

\label{tab:results}
\end{table}

\vspace{-20pt}

 \subsection{Results and Analysis}

Two variations of our model were evaluated: the Diffusion-based MaTCHS and the naive MaTCHS, which omits diffusion. The naive MaTCHS uses an input size of $N\times P \times L$ for $N$ stocks, $P$ indicators, and $L$ timesteps, with an adjusted output layer ($N\times1$) for next-day stock predictions. Comparative results are presented in Table-1 and Table-2.

The Diffusion-based MaTCHS excels on the StockNet Dataset\cite{xu-cohen-2018-stock}, outperforming all other models in F1, accuracy, and MCC metrics, and on NASDAQ and NYSE dataset\cite{feng2019temporal} outperforming others on SR and IRR . Without the diffusion component, MaTCHS still performs admirably, matching the STHAN-SR model's performance. This emphasizes the strength of our Masked Relational Transformers (MRT) in grasping complex inter-stock dynamics over other GNN-based techniques like HyperGraph-structured STHAN-SR. Separating temporal and spatial predictions has proved beneficial, addressing the complexities arising from concurrent modeling. This separation fosters precision and leads to superior price trend forecasting.

\textbf{Our Diffusion architectures'} performance underscores our hypothesis on the diffusion models ability in capturing stock market nuances better due to their probabilistic nature. Further, our specialized noise schedule for Diffusion enhances performance over standard Diffusion noise schemes across all Datasets. Our guidance in Diffusion emphasizes learning volatile and relational trends, hence augmenting the denoising models' capability. 

Further NASDAQ and NYSE datasets, our models reveal potent Portfolio Return trends. Notably, the Sharpe ratios of 7.92\% and 6.18\% suggesting considerable advances over previous models for utilisation for automated trading capablities. Also, higher IRR scores signify the model's ability to incorporate distant temporal dynamics in prediction, as higher IRR indicates better annual returns relative to the amount invested.

We also trained and tested CSDI\cite{tashiro2021csdi} and TimeGrad\cite{rasul2021autoregressive} on StockNet(Table-3), we noted predictable performance declines. As CSDI focuses on time-series imputation, and TimeGrad's diffusion-based training isn't tailored to our goals. Our CRPS scores outperformed both, proving our diffusion model's superiority in capturing data distribution.

An ablation study, examining the impact of aggregating relations across attention heads, exhibited performance drops across the StockNet, NASDAQ, and NYSE datasets. Amplifying the attention heads number yields marginal improvements, reinforcing the rationale behind the MRT's design.

\vspace{-10pt}
\subsection{Limitations and Conclusion}

Our architecture, while effective, requires substantially more computational resources and time compared to alternatives, hindering its applicability in rapid scenarios like day-trading. Immediate reductions in iterations compromise model efficacy. More efficient diffusion architectures are essential for real-world use. In summary, our diffusion-based stock prediction architecture outperforms current models, presenting a promising avenue for improved stock market predictions and advancing research in this sector.

\bibliographystyle{IEEEbib}
\bibliography{strings,refs}

\end{document}